\documentclass[11pt]{article}

\usepackage[preprint]{acl}

\usepackage{times}
\usepackage{latexsym}

\usepackage[T1]{fontenc}

\usepackage[utf8]{inputenc}

\usepackage{microtype}

\usepackage{inconsolata}

\usepackage{graphicx}

\usepackage{amssymb}
\usepackage{amsmath}
\usepackage{algorithm}
\usepackage{algorithmic}
\usepackage{cleveref}
\usepackage{graphicx}
\usepackage{bbding}
\usepackage{pifont}
\usepackage{multirow}
\usepackage{lineno}
\usepackage{booktabs}
\usepackage{tabularx}
\usepackage{arydshln}
\usepackage{listings}

\usepackage{times} 
\usepackage{helvet} 
\usepackage{courier} 
\usepackage{natbib}  
\usepackage{caption} 

\usepackage[most]{tcolorbox}
\tcbset{colframe=black!50, colback=gray!5, fonttitle=\bfseries}

\title{LLM Agents for Deliberative Collaboration: A Study on \\ Joint Decision Making Under Partial Observability}

\author{
    Chenxu Wang\textsuperscript{*,1},
    Yongkun Yang\textsuperscript{*,1},
    Boyuan Du\textsuperscript{*,2},
    Shiwei Lin\textsuperscript{1},
    Huaping Liu\textsuperscript{1}
    \\
    \textsuperscript{1}Department of Computer Science and Technology, Tsinghua University
    \\
    \textsuperscript{2}Fuzhou University
    \\
    \small{
    \href{mailto:jimwangcx@gmail.com}{jimwangcx@gmail.com},
    \href{mailto:hpliu@tsinghua.edu.cn}{hpliu@tsinghua.edu.cn}
    }
}

\def\ie{\emph{i.e.}}

\lstdefinestyle{promptstyle}{
    basicstyle=\rmfamily,
    breaklines=true,
    columns=fullflexible,
    frame=b,
    framerule=0.8pt,
    aboveskip=0.5em,
    belowskip=0pt,
}

\lstdefinestyle{smallpromptstyle}{
    basicstyle=\small\rmfamily,
    breaklines=true,
    columns=fullflexible,
    frame=b,
    framerule=0.8pt,
    aboveskip=0.5em,
    belowskip=0pt,
}

\lstdefinestyle{jsonstyle}{
    basicstyle=\small\ttfamily,
    breaklines=true,
    columns=fullflexible,
    frame=single,
    framerule=0.8pt,
    aboveskip=0.5em,
    belowskip=0.5em,
}

\lstdefinestyle{textblockstyle}{
    basicstyle=\footnotesize\ttfamily,
    breaklines=true,
    columns=fullflexible,
    frame=single,
    framerule=0.8pt,
    aboveskip=0.5em,
    belowskip=0.5em,
}

\begin{document}
\maketitle

\begingroup
\renewcommand{\thefootnote}{\fnsymbol{footnote}}
\footnotetext[1]{Equal contribution.}
\endgroup

\begin{abstract}
Deliberation plays a crucial role in collaboration; when humans work together, they naturally engage in communication to align information and reach an agreement. In this paper, we investigate deliberative large language model (LLM) agents under partially observable joint decision-making tasks.
We formalize deliberative collaboration as a cooperative joint decision problem with partial and asymmetric observations, and introduce a scalable benchmark that instantiates this problem across multiple task settings and domains in which agents must exchange information through deliberation to reach a joint decision with a shared reward.
We then instantiate a reference scaffold and evaluation protocol for deliberative agents and conduct a systematic evaluation of a range of representative LLMs.
The results reveal that complex deliberative collaboration tasks continue to challenge state-of-the-art language models. Even with the aid of external mathematical tools, language models may fail in either the deliberation process for aligning information or the complex reasoning process for making the decision. 
On the other hand, diagnostic analysis reveals that the deliberation process may also provide opportunities for reflection and error correction, sometimes improving performance over centralized baselines.
Altogether, our work establishes a foundation for evaluating and improving LLM agents in deliberative collaboration and provides insights into the strengths, limitations, and properties of current LLM-based multi-agent systems. 
Code is available at \href{https://github.com/wcx21/deliberative-collaboration-agents}{this repository}.

\end{abstract}

\section{Introduction}
\begin{figure}[t]
\centering
\includegraphics[width=\linewidth]{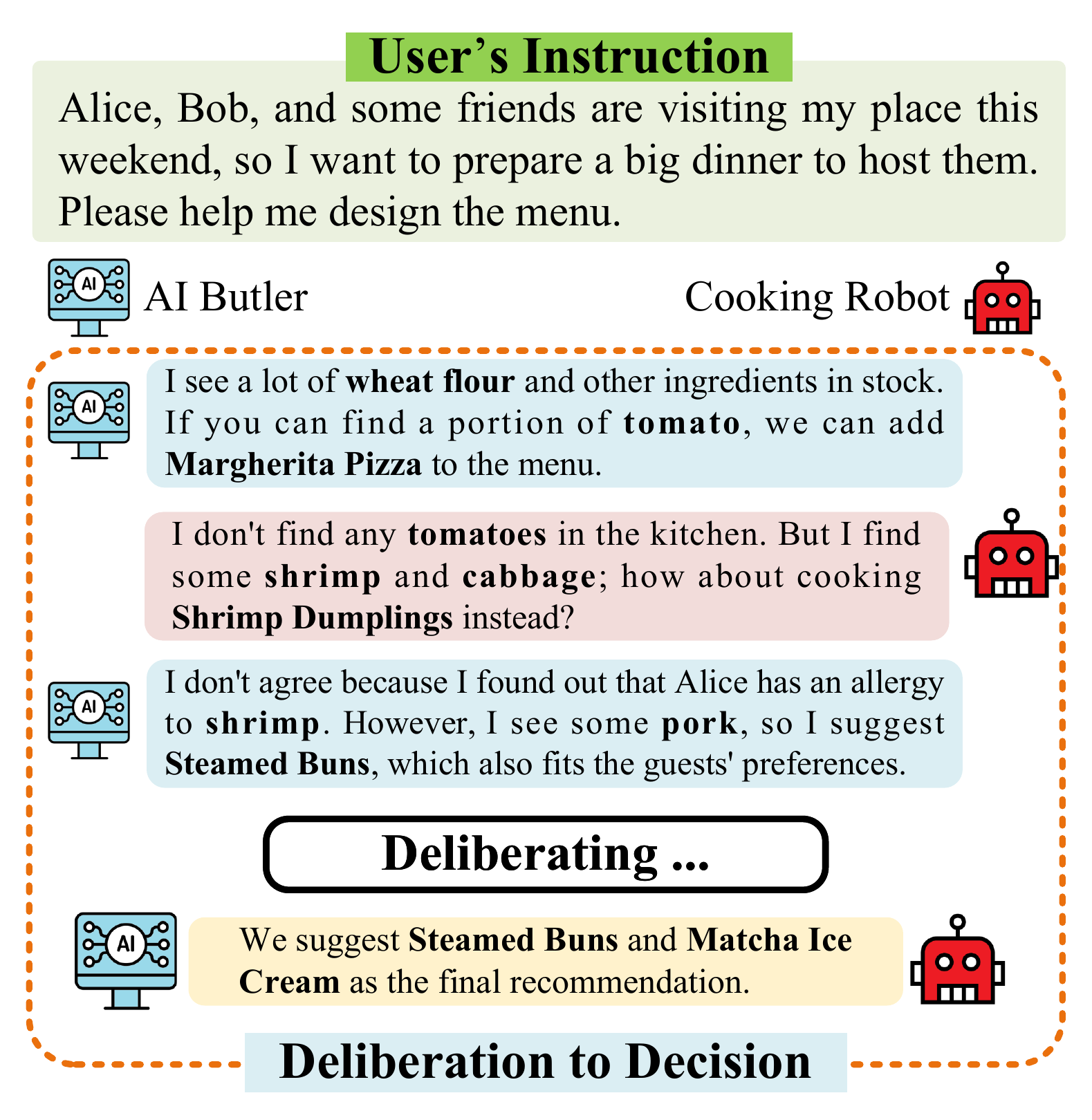}
\caption{An illustrative example of deliberative collaboration under partial observability. In this menu design scenario, multiple AI agents (e.g., an AI butler and a cooking robot) observe different aspects of the environment, such as ingredient availability and guest preferences, and must exchange information through deliberation to reach a single joint decision. This structure can generalize to other cooperative joint decision problems.}

  \label{fig: story}
\end{figure}

Recent advances in large language models (LLMs) and language agents have achieved impressive progress across a variety of domains, including social intelligence \cite{zhou2025socialeval, mou2025agentsense, wang2024towards}, multi-agent collaboration \cite{yan2025beyond, chen2025optima, guo2024large, liu2024capo}, negotiation or deliberation \cite{zhu2025automated, choi2025debate,hu2025debate, bianchi2024well,  karanam2024towards}, and decision making \cite{dolant2025agentic, eo2025debate}.
However, an open question remains: can LLM agents collaborate well through deliberation when they each possess only partial, asymmetric information? 
We present a representative example of deliberative collaboration in \Cref{fig: story}. In deliberative collaboration scenarios, the agents must communicate, exchange their knowledge, and reason together to reach a shared decision. 
This setting is inherently more complex and comprehensive than pure deliberation or collaboration, as it demands task-related communication and joint decision-making.

In this paper, we study LLM agents in deliberative collaboration scenarios. We formalize deliberative collaboration as a partially observable joint decision-making problem, where multiple agents each have their own partial observation of the environment, and the agents should reach a consensus on a final decision as the output and share the same reward. 
Unlike competitive or adversarial settings, the challenge in such scenarios lies in effective information exchange, belief alignment, and coordinated decision-making under partial observability.

We build a scalable benchmark that instantiates this problem across multiple related task settings, across collaborative menu design and task allocation domains. The settings differ in observation structure, agent role, and decision mechanism, while sharing the same underlying deliberative collaboration structure. 
The underlying decision problem is formulated as a combinatorial optimization problem with finite decision spaces, enabling rule-based numerical rewards.
The benchmark enables a comprehensive evaluation of LLM agents’ capabilities in deliberative collaboration, including information exchange through conversation, numerical calculations, and logical reasoning. Although instantiated in concrete domains, the underlying problem structure, cooperative joint decision making with incomplete and asymmetric information, is shared by many real-world applications.

We conduct experiments with multiple state-of-the-art LLMs on 180 tasks across a range of task settings and evaluation conditions. In particular, we inspect the effects of the decentralized deliberation process, external tools, and model capacity on collaborative decision making. 
Results reveal that current models may still have shortcomings in information exchange and aggregation, reasoning, and math problem solving. However, while challenging various capabilities of the LLMs, the deliberation process also provides a chance for reflection and error correction for the agents and may improve the performance in specific circumstances.

Our main contributions are summarized as follows:
\begin{itemize}
    \item We formalize deliberative collaboration as a collaborative joint decision-making problem under partial observations, providing a unified abstraction for studying cooperative LLM agents under incomplete and asymmetric information.
    \item We introduce a scalable benchmark that instantiates this problem across multiple tightly related task settings, including menu design and task allocation, enabling controlled evaluation under various observation and decision structures with solver-computable objectives.
    \item We conduct a systematic evaluation across multiple LLMs and task settings, establishing baseline results and diagnostic observations for current deliberative agent systems.
\end{itemize}

\section{Benchmarking Joint Decision Making Under Partial Observability}
This work is motivated by a class of real-world collaboration scenarios, where multiple agents cooperate to solve a task. The agents may have distinct observations and action spaces while sharing the same reward; therefore, to achieve a higher common reward, deliberation would be a natural requirement.
We conduct a study on a representative set of deliberative collaboration scenarios that can be formulated as a decision problem under partial observability.

\subsection{Problem Formulation}
\label{sec: problem_def}
We define the partial observable multi-agent joint decision problem as a tuple: $(s, O_{1}, \cdots, O_{n}, \mathcal{D}, R)$, where $s$ is the ground truth state in the problem which cannot be observed by the agents; $O_{1}, \cdots, O_{n}$ are the partial observations of the agents, where $n$ is the number of agents; $\mathcal{D}$ is the decision space; and $R: \mathcal{S} \times \mathcal{D} \mapsto \mathbb{R}$ is the reward function, where $\mathcal{S}$ is the state space. The goal of the agents is to select a decision $d \in \mathcal{D}$ that maximizes the reward $R(s, d)$, with respect to the unobserved ground truth state $s$. 
To enable the agents to reach a joint decision, we associate the problem with a deliberation process, as described in a programming-friendly manner through the pseudocode in \Cref{algo: deliberation_process}, where agents are defined in a general manner and \textit{decision\_made} is a customizable function that determines whether the agents reached an agreement. The protocol is intentionally abstract to support customized features; for example, role asymmetry can be instantiated by restricting which agents are taken into account by the \textit{decision\_made(·)} function.

Note that although the deliberation process unfolds over multiple turns, the utterances serve solely to exchange information, and the environment state is fixed, so the underlying task is a single-stage joint decision problem with a terminal reward. While this setting could be formulated in a Dec-POMDP manner, our formulation underlines the task's nature as a one-shot collective decision problem and focuses more on the deliberation process as epistemic coordination instead of the dialogue as state transition.

\begin{algorithm}[t]
    \caption{Deliberation Process}
    \begin{algorithmic}
        \label{algo: deliberation_process}
        \STATE {\textbf{Input}: Observations $O_{1}, \cdots, O_{n}$, Language Agents $\pi_{1}, \cdots, \pi_{n}$, Decision Space $\mathcal{D}$, Turn limitation $t_m$}
        \FOR{$i \in \{1, \cdots, n\}$}
            \STATE {$\pi_{i}\text{.init}(O_{i})$}
        \ENDFOR
        
        \STATE {$t \gets 0$}
        \STATE {$dialog \gets \emptyset$}
        \WHILE{t $\leq$ $t_m$}
            \FOR{$i \in \{1, \cdots, n\}$}
                \STATE {$\pi_{i}\text{.listen}(dialog)$}
                \STATE {$\text{utterance, current\_proposal} \gets \pi_{i}\text{.talk}()$}
                \STATE {$dialog \gets dialog.append(utterance)$}
                \IF{decision\_made($\cdot$)}
                    \STATE {\textbf{Return} \text{current\_proposal}}
                \ENDIF
            \ENDFOR  
            \STATE {$t \gets t + 1$}
        \ENDWHILE
        \STATE {\textbf{Return} None}
    \end{algorithmic}
\end{algorithm}

\subsection{Benchmark Design}
\label{sec: benchmark_designation}

We instantiate the partial observable joint decision problem in a set of task settings that vary in observation structure, decision authority, and evaluation protocol. Across all settings, agents must reach a single joint decision through deliberation based on partial observations, while sharing a common terminal reward. The key differences lie in the form of observations, the structure of the decision space, and the evaluation protocol, as summarized in \Cref{tab: benchmark}.

\subsubsection{Menu Design with Partitioned Numerical Observations}

We start with a two-agent menu design scenario, with a setting that instantiates the joint decision problem in a fully structured environment with numerical state representations and partitioned observations. 
The ground truth state consists of the ingredient state and user preferences, both of which can be represented in the form of vectors; the agents observe disjoint subsets of this state, with the guarantee that their observations jointly cover the full information. 
The decision space is a finite set of candidate menus. Given a dish list, for each dish, the agents decide whether to include it or not in the menu, and the reward is deterministically computed based on feasibility and user preference scores. Given access to the full state, the optimal decision can be obtained by solving a corresponding integer programming problem, which serves as an upper bound. As a result, this setting primarily evaluates agents’ ability to exchange essential information through deliberation and to perform accurate planning once sufficient information has been aligned. 
More details are elaborated in Appendix \ref{sec: menu_design_appendix}.

\begin{table*}[t]
\centering

\begin{tabular}{lcccc} 
\hline
Task Setting & Observation Form & Observation Coverage & Agent Role & Decision Authority   \\  
\hline
Menu–Numeric & Numerical & Partitioned & Symmetric & Symmetric \\ 
Menu–Semantic & Hybrid & Partial  & Symmetric & Symmetric \\
Task Allocation & Numerical & Partitioned & Asymmetric & Leader-based \\ 

\hline
\end{tabular}

\caption{Comparison of task settings instantiated in the benchmark. Hybrid observations include both numerical information and natural language descriptions.}
\label{tab: benchmark}
\end{table*}

\subsubsection{Menu Design with Partial Semantic Observations}

Considering that real-world observations may be textual and semantic instead of numerical, we relax the assumption that observations are purely numerical and disjoint. In this setting, agents receive partial natural language descriptions of the user preferences, where the natural language descriptions are sampled from the ground truth state and may be incomplete, overlapping, or correlated, thus different agents may describe the same latent fact from distinct perspectives.

The decision space and reward function remain the same as in the structured menu design setting, but agents must now reason under uncertainty induced by language-mediated observations; therefore, even an oracle baseline that has access to all natural language observations may not achieve the full reward. This setting emphasizes natural language level reasoning and communication, interpretation of ambiguous information, and deliberation under partial observability, while preserving a well-defined objective for evaluation.

\subsubsection{Task Allocation with Asymmetric Roles and Resources}

We extend the joint decision framework to a multi-agent task allocation scenario characterized by asymmetric roles and heterogeneous resource constraints. Compared to the menu design domain, resources in this domain are divided into private resources (exclusive to individual agents) and public resources (shared across the team), together with agent-specific efficiency for tasks. The partial observability is also associated with this feature, where each agent can only observe its own private resources and a part of the public resources, requiring communication through deliberation to reconstruct global feasibility constraints.
The decision space consists of assigning tasks to agents under mixed private and public resource constraints, where each task can be assigned to at most one agent. Within the resource constraints, the reward is calculated according to the base value of each task and the agents' efficiency in performing the task. Akin to the menu design domain, if the ground truth state is known, the task allocation problem can also be formulated as a combinatorial optimization problem and has a computable upper bound of reward calculated by the solver.

In contrast to the symmetric peer interaction in menu design, this setting introduces explicit role asymmetry. One agent is designated as the leader agent, while others are worker agents. All agents can communicate to exchange information and make a proposal, but only the leader agent can make the final decision and terminate deliberation.
This setting differs fundamentally from menu design in both decision structure and authority distribution, and serves to evaluate agents’ ability to perform hierarchical coordination under partial observability.

\subsubsection{Metrics}

All domains in our benchmark are associated with a well-defined structured decision space and objective reward, thus we can evaluate agents using normalized reward (\textbf{NR} $ = \frac{\text{reward}}{\text{maximum reward}} \times 100$), defined as the achieved reward divided by the maximum possible reward under the ground truth state. 
Invalid decisions that violate feasibility constraints receive zero reward for NR. Therefore, we also report the valid ratio (\textbf{VR}), defined as the proportion of valid decisions, \ie, satisfying all feasibility constraints.

To provide a more fine-grained assessment for diagnosis, we provide additional metrics in \Cref{sec: appendix_Metric_Details}. Together, these metrics capture both optimality and constraint adherence under partial observability.

\subsubsection{External Tools}

To disentangle the challenge of mathematical calculation from the deliberation process, we provide two optional symbolic tools that agents can invoke during the planning phase. 
The Solver formulates the decision problem as an integer linear programming instance, deriving the utility-maximizing solution based solely on the agent's current estimated belief of the global state.
Complementing this, the Calculator serves as a fine-grained feasibility oracle that verifies constraints ranging from simple ingredient availability to hierarchical private-public resource limits, while returning diagnostic feedback on specific resource deficiencies for invalid proposals.
This framework adapts to the specific constraint structures of each domain: menu design is modeled as subset selection against a single shared inventory, whereas task allocation enforces a hierarchical system involving both local private capacities and global public budgets.

\subsection{Benchmark Implementation}
\label{sec: benchmark_implementation}

We implement the benchmark using a database-driven generation pipeline to ensure scalability, diversity, and reproducibility across task settings. 
The benchmark consists of two databases: a menu design database for \textit{menu-numeric} and \textit{menu-semantic}, and a teamwork database for \textit{task allocation}. Following prior works \cite{zhou2024sotopia, wang2024dynamic, wang2024towards}, the databases are initially generated using LLMs and subsequently reviewed and refined by the authors to ensure consistency and commonsense validity. Then we sample agent personas, states, and observations from the database to generate task instances for each domain.
Further details are in \Cref{sec: benchmark_appendix}.

The menu design database contains 51 ingredients, 78 dishes spanning 12 cuisines, and 10 character profiles representing guests. Each character is defined by a set of attributes that may influence food preferences, including demographic information, dietary constraints, cultural background, health goals, and dining style. Each menu design task instance is generated by sampling an ingredient inventory, a subset of candidate dishes with corresponding recipes, and two guest characters. This procedure applies to both \textit{menu-numeric} and \textit{menu-semantic}, which only differ in the observation sampling.
In total, we generate 60 instances for each menu design setting.

For task allocation, we construct the database across three sub-domains: research project collaboration, cafeteria operations, and camping in the wild. The database contains a total of 42 task templates, 12 private resource types, 12 public resource types, and 18 agent personas. 
Each task allocation instance samples 10 tasks from the domain task pool and three agents (one as the leader and two as workers) from the persona pool. Resources and task values are sampled from persona-specific distributions. 
We generate 20 instances per sub-domain, for a total of 60.

\section{Deliberative Agent}
\label{sec: agent_design}

\begin{figure}
    \centering
    \includegraphics[width=0.9\linewidth]{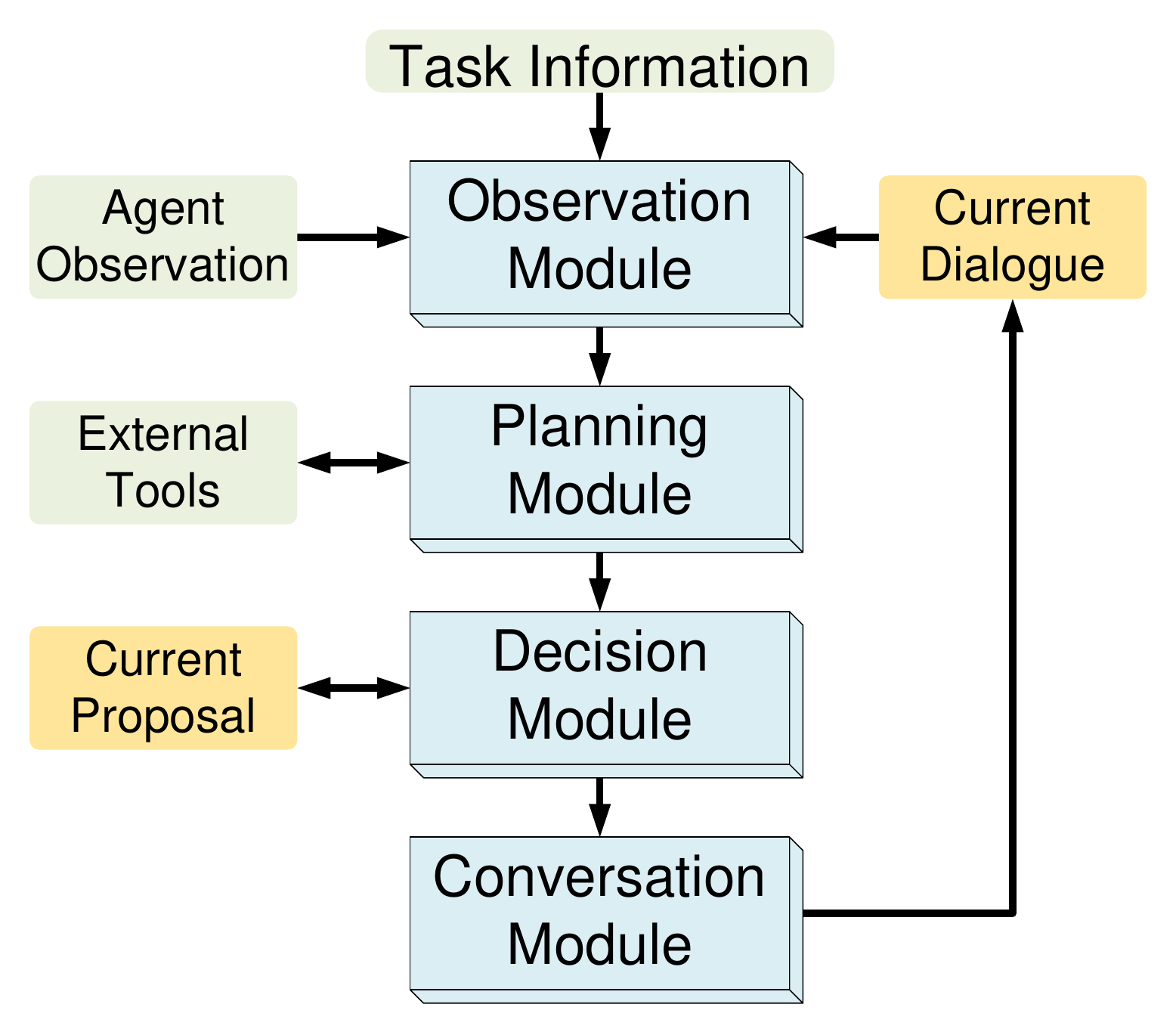}
    \caption{Illustration of our agent framework for solving the deliberative joint decision-making tasks. 
    Green blocks indicate inputs, blue blocks represent agent modules, and orange blocks show variables modified by the agent.
    }
      \label{fig: agent}
\end{figure}

To construct LLM agents for deliberative collaboration and instantiate the evaluation, we introduce an agent framework as a fixed reference scaffold, as shown in \Cref{fig: agent}, which includes modules of observation, planning, decision, and conversation.
As a preliminary approach, we implement task-specific agents to solve the deliberative joint decision problems. Agents across all domains share the same general architecture, while the prompts and implementation are slightly different to fit each task setting.
Specifically, the \textbf{observation module} takes the initial observation and the dialogue as inputs to estimate the ground truth state, 
covering domain-specific factors such as ingredients and dish values, or private resource distribution and agent efficiencies. After that, the \textbf{planning module} proposes a candidate menu or task allocation scheme based on the estimated values, along with supporting reasons. If external tools are available, the module will get the result returned by the tools as part of the input. Then the \textbf{decision module} decides whether to accept the current proposal based on all information up to now. If the agent declines the current proposal, the \textbf{conversation module} will be called to generate a message to talk to the partner. For the task allocation domain, the leader agent also decides whether to terminate the process and submit the final decision.
General task information and instructions are used throughout the whole procedure. 
In each module, the LLM is prompted to have a thinking process in the output content to implement a chain of thought \cite{wei2022chain}.

\begin{table*}[t]
\centering
\setlength{\tabcolsep}{3.5pt}
\begin{tabular}{l c cc cc cc}
\hline
\multirow{2}{*}{Model} &
\multirow{2}{*}{Tools} &
\multicolumn{2}{c}{Menu-Numeric} &
\multicolumn{2}{c}{Menu-Semantic} &
\multicolumn{2}{c}{Task Allocation} \\
\cline{3-8}
& & NR (mean$\pm$se) & VR(\%) & NR (mean$\pm$se) & VR(\%) & NR (mean$\pm$se) & VR(\%) \\
\hline
\multicolumn{8}{c}{\textit{Deliberative Collaboration}} \\
\hline

GPT-5.1 & \ding{55} & 95.60$\pm$1.51 & 100.00 & 55.95$\pm$6.25 & 60.00  & 80.28$\pm$2.75 & 98.33 \\
        & \ding{51} & 90.31$\pm$3.02 & 98.33  & 54.60$\pm$5.75 & 66.67  & 76.06$\pm$3.11 & 96.67 \\

DeepSeek-V3.2 & \ding{55} & 88.36$\pm$3.68 & 91.67 & 78.46$\pm$3.96 & 93.33 & 53.45$\pm$5.61 & 61.67 \\
              & \ding{51} & 95.66$\pm$1.93 & 98.33 & 83.90$\pm$2.34 & 100.00 & 80.59$\pm$3.96 & 90.00 \\

GLM-4.7 & \ding{55} & 91.55$\pm$2.35 & 98.33 & 65.03$\pm$3.58 & 95.00 & 60.09$\pm$4.95 & 73.33 \\
        & \ding{51} & 86.96$\pm$3.58 & 93.33 & 64.65$\pm$3.67 & 93.33 & 64.36$\pm$4.13 & 83.33 \\

GPT-4.1-mini & \ding{55} & 12.07$\pm$4.05 & 13.33 & 28.20$\pm$5.26 & 35.00 & 49.70$\pm$5.03 & 65.00 \\
             & \ding{51} & 35.53$\pm$5.91 & 38.33 & 24.17$\pm$4.99 & 30.00 & 67.98$\pm$4.69 & 80.00 \\

Qwen3-Next-80B & \ding{55} & 34.63$\pm$5.60 & 41.67 & 38.34$\pm$4.09 & 75.00 & 53.09$\pm$3.68 & 83.33 \\
               & \ding{51} & 25.04$\pm$5.37 & 28.33 & 47.99$\pm$4.65 & 76.67 & 52.54$\pm$3.61 & 83.33 \\

Qwen3-32B & \ding{55} & 4.78$\pm$2.41 & 6.67  & 9.75$\pm$2.98  & 18.33 & 6.04$\pm$2.26  & 11.67 \\
          & \ding{51} & 18.77$\pm$4.60 & 25.00 & 15.87$\pm$3.85 & 28.33 & 47.31$\pm$5.88 & 55.00 \\

Qwen3-30B & \ding{55} & 28.64$\pm$5.22 & 36.67 & 30.10$\pm$4.46 & 50.00 & 55.15$\pm$2.95 & 90.00 \\
          & \ding{51} & 29.27$\pm$5.36 & 36.67 & 33.94$\pm$4.63 & 53.33 & 60.04$\pm$3.20 & 91.67 \\

\hline
\multicolumn{8}{c}{\textit{Centralized Baseline}} \\
\hline

GPT-5.1 & \ding{55} & 96.71$\pm$1.22 & 100.00 & 90.67$\pm$1.37 & 100.00 & 91.67$\pm$0.95 & 100.00 \\
        & \ding{51} & 98.42$\pm$0.80 & 100.00 & 90.84$\pm$1.46 & 100.00 & 93.41$\pm$1.87 & 98.33 \\

DeepSeek-V3.2 & \ding{55} & 97.36$\pm$1.00 & 100.00 & 92.46$\pm$1.35 & 100.00 & 79.82$\pm$3.93 & 88.33 \\
              & \ding{51} & 98.14$\pm$0.84 & 100.00 & 92.28$\pm$1.21 & 100.00 & 89.33$\pm$3.89 & 90.00 \\

GLM-4.7 & \ding{55} & 95.66$\pm$1.88 & 98.33 & 75.58$\pm$3.40 & 93.33 & 58.51$\pm$5.47 & 66.67 \\
        & \ding{51} & 97.46$\pm$1.75 & 98.33 & 86.49$\pm$2.42 & 98.33 & 63.39$\pm$5.50 & 70.00 \\

GPT-4.1-mini & \ding{55} & 94.64$\pm$1.25 & 100.00 & 84.64$\pm$2.48 & 98.33 & 40.53$\pm$4.82 & 56.67 \\
             & \ding{51} & 99.26$\pm$0.43 & 100.00 & 87.88$\pm$1.72 & 100.00 & 44.79$\pm$6.12 & 48.33 \\

Qwen3-Next-80B & \ding{55} & 90.02$\pm$3.05 & 95.00 & 78.10$\pm$2.86 & 96.67 & 76.91$\pm$4.10 & 86.67 \\
               & \ding{51} & 95.24$\pm$2.42 & 96.67 & 80.83$\pm$2.54 & 98.33 & 69.58$\pm$5.53 & 73.33 \\

Qwen3-32B & \ding{55} & 10.85$\pm$3.68 & 13.33 & 18.05$\pm$4.59 & 21.67 & 1.11$\pm$1.11 & 1.67 \\
          & \ding{51} & 90.88$\pm$3.60 & 91.67 & 69.31$\pm$5.17 & 76.67 & 28.21$\pm$5.84 & 28.33 \\

Qwen3-30B & \ding{55} & 91.54$\pm$3.35 & 93.33 & 81.33$\pm$2.50 & 98.33 & 57.00$\pm$5.38 & 66.67 \\
          & \ding{51} & 98.45$\pm$0.71 & 100.00 & 77.16$\pm$2.62 & 98.33 & 36.20$\pm$5.84 & 40.00 \\

\hline
\end{tabular}

\caption{Overall results across domains. We report normalized reward (mean$\pm$standard error) and valid ratio (\%).}
\label{tab: main_results}
\end{table*}

\section{Experiments}

\subsection{Experimental Setup}
Since our benchmark is purely cooperative, we evaluate the language agents in a self-play manner. We select a combination of representative cutting-edge models and affordable models, including GPT-5.1~\footnote{gpt-5.1-2025-11-13}, DeepSeek-V3.2~\cite{liu2025deepseek}, GLM-4.7~\cite{glm47_blog}, GPT-4.1-mini~\footnote{gpt-4.1-mini-2025-04-14
}, Qwen3-Next-80B-A3B (abbreviated as \textbf{Qwen3-Next-80B}), Qwen3-32B, and Qwen3-30B-A3B (abbreviated as \textbf{Qwen3-30B}),~\cite{yang2025qwen3}. For a fair comparison in reasoning effort and token usage, we disable the internal thinking for all models.

To balance conversation diversity and result stability, we set all LLM temperatures to 0.3. 
The deliberation process lasts for at most 6 rounds.
For a comprehensive evaluation, we conduct experiments under both conditions with and without tools. 
Additionally, to study the challenges posed by the decentralized system model, we introduce a \textbf{centralized baseline} in which a centralized agent receives all perfectly merged observations and makes a single-round decision.

\subsection{Main Results}
We present primary experiment results including NR and VR across all three domains in \Cref{tab: main_results}.
Cutting-edge models, including GPT-5.1, DeepSeek-V3.2, and GLM-4.7, achieve significantly higher rewards than smaller models such as GPT-4.1-mini and Qwen3-30B. The large models achieve normalized rewards over 90 in the \textit{menu-numeric} domain, indicating that deliberative collaboration in simple settings can be adequately solved by the state-of-the-art LLMs, though it may still be challenging for small models. However, when the task gets complex, such as in the \textit{task allocation} domain, there is still room for improvement even for large models.

Comparing across domains, all models tend to achieve the highest rewards in \textit{menu-numeric}, lower in \textit{menu-semantic} and \textit{task allocation}. This is in line with expectations since the \textit{menu-semantic} introduces uncertainty and reasoning hardness into the task, and \textit{task allocation} has a much more complex math problem; both are more difficult than \textit{menu-numeric}.

\subsection{Centralization vs. Deliberation}

\begin{table*}[t]
\centering
\begin{tabular}{l c ccc ccc}
\hline
\multirow{2}{*}{Model} & \multirow{2}{*}{Tools}
& \multicolumn{3}{c}{NR}
& \multicolumn{3}{c}{VR(\%)} \\
\cline{3-8}
& & Oracle &   $\Delta_{\text{O-Cent.}}$ &   $\Delta_{\text{O-Delib.}}$
  & Oracle &   $\Delta_{\text{O-Cent.}}$ &   $\Delta_{\text{O-Delib.}}$ \\

\hline
GPT-5.1
    & \ding{55} & 95.54$\pm$0.61 & +3.87  & +15.26 & 100.00 & +0.00  & +1.67 \\
    & \ding{51} & 99.08$\pm$0.47 & +5.67  & +23.02 & 100.00 & +1.67  & +3.33 \\

DeepSeek-V3.2
    & \ding{55} & 87.78$\pm$2.82 & +7.96  & +34.33 & 95.00  & +6.67  & +33.33 \\
    & \ding{51} & 93.87$\pm$2.84 & +4.54  & +13.28 & 95.00  & +5.00  & +5.00 \\

GLM-4.7
    & \ding{55} & 79.50$\pm$4.17 & +20.99 & +19.41 & 86.67  & +20.00 & +13.34 \\
    & \ding{51} & 81.54$\pm$4.83 & +18.15 & +17.18 & 88.33  & +18.33 & +5.00 \\

GPT-4.1-mini
    & \ding{55} & 55.01$\pm$4.73 & +14.48 & +5.31  & 71.67  & +15.00 & +6.67 \\
    & \ding{51} & 96.32$\pm$1.86 & +51.53 & +28.34 & 98.33  & +50.00 & +18.33 \\
\hline
\end{tabular}
\caption{The performance of the oracle baseline in the task allocation domain. $\Delta_{\text{O-Cent.}}$ and $\Delta_{\text{O-Delib.}}$ denote the NR increase of the oracle baseline compared to the centralized baseline and standard deliberative collaboration setting, respectively.}
\label{tab:oracle_task_allocation}
\end{table*}

\begin{table*}[t]
\centering
\begin{tabular}{l ccc ccc}
\hline
\multirow{2}{*}{Model} &
\multicolumn{3}{c}{Menu-Numeric} &
\multicolumn{3}{c}{Task Allocation} \\
\cline{2-7}
& Tool NR & Actual NR & Comp. (\%) & Tool NR & Actual NR & Comp. (\%) \\
\hline
GPT-5.1         & 94.22 & 93.09 & 86.21 & 96.67 & 76.06 & 0.00 \\
GLM-4.7         & 94.22 & 89.81 & 71.19 & 78.06 & 64.36 & 12.28 \\
Qwen3-Next-80B  & 42.26 & 25.32 & 27.59 & 58.56 & 52.54 & 22.03 \\
\hline
\end{tabular}
\caption{Additional indicators of tool use in selected scenarios. \emph{Tool NR} denotes the NR of the solution provided by the tool in the last round. \emph{Actual NR} denotes the NR of the model's final decision. \emph{Comp. (\%)} denotes the compliance rate, i.e., the proportion of cases where the model's final decision exactly matches the final tool output.}
\label{tab:tool_compliance_main}
\end{table*}

One may expect agents in the centralized baseline to perform better than in the standard decentralized setting, since the agent can directly get the perfectly merged observation without information loss or hallucination. However, we find that the effect of deliberation is complex and subtle in practice. Though agents get a significant gain in reward in most scenarios, in some specific cases, such as GLM-4.7 and GPT-4.1-mini in the \textit{task allocation} domain, the performance even reduces.

The performance gain in the centralized baseline may partly come from more accurate state estimation.
Additional diagnostics in \Cref{sec: appendix_exp} show that state-estimation errors, measured by hallucination rate or NMAE, are associated with performance gaps in some settings.

As for the unusual performance drop, after qualitative inspection of the relevant cases, we empirically find that the deliberation process provides a chance for reflection and error correction. Therefore, it is possible for the centralized baseline to achieve a lower reward since it does not have the deliberation process. An example and case study is provided in \Cref{sec:appendix_oracle_case_study}.

To further diagnose the roles of information access and multi-round discussion in the deliberation process, we introduce an \emph{oracle} baseline that combines the deliberation process with perfectly merged full observations.
We conduct an additional experiment with the most representative models on the \textit{task allocation} domain, where the performance drop happened, and present the results in \Cref{tab:oracle_task_allocation}. 
As shown by the results, the oracle baseline exhibits further performance gain in all cases. For GLM-4.7 and GPT-4.1-mini, the oracle baseline significantly outperforms the centralized baseline, whereas the performance gap is not so large for GPT-5.1 and DeepSeek-V3.2. Such evidence supports the beneficial effects of the deliberation process for LLM agents in making decisions and explains the unexpectedly low performance of GLM-4.7 and GPT-4.1-mini in the centralized baseline.

\subsection{Effect of External Tools}

In most scenarios, external mathematical tools substantially improve the performance of the agent, especially for DeepSeek-V3.2, GPT-4.1-mini, and Qwen3-32B. This adheres to the current consensus that LLMs may not be directly good at calculation, and mathematical tools can help.
However, the effect of external tools seems to be model-specific, and there are even some counterintuitive results. For example, GPT-5.1, GLM-4.7, and Qwen3-Next-80B do not gain much more reward with the help of tools; the reward even reduces in some cases.

To explain this phenomenon, we manually inspect the counterintuitive cases and conduct an additional process analysis for scenarios where counterintuitive cases happen. 
We present additional indicators in representative scenarios in \Cref{tab:tool_compliance_main}. 
Although the mathematical tools provide a good solution, the agents tend not to adopt the solution but try to do the calculation themselves and propose another solution instead. GPT-5.1 even shows a compliance rate of 0 in the task allocation domain.
However, due to limited computational capability, the LLMs may not obtain the optimal solution. We provide an example in \Cref{sec:appendix_skepticism_trap_case_study} as a case study.
Another factor that undermines the usefulness of external tools is the accumulated error in state estimation. Since the solver calculates the best solution based on the input state, for some less intelligent models, if the state estimation from the partial observation is inaccurate, the solver will produce a wrong solution as well.

\section{Related Work}

\subsection{Language Agents and Multi-Agent Systems}
With advances in LLM techniques and their achievement of near-human performance in many aspects, a large body of prior work has shown the feasibility of building LLM-based autonomous agents to solve complicated tasks, known as language agents \cite{wang2024survey, liua2024gentbench,zhu2025multiagentbench, ferrag2025llm}.

Language agents can also form multi-agent systems \cite{tran2025multi}, which can not only be used as a technology to improve the performance of LLM-based systems \cite{ chen2024internet, qian2024chatdev,wang2025mac,he2025llm}, but also can apply to intrinsically multi-agent, such as social simulations and social intelligence \cite{wang2024towards, zhou2025socialeval, mou2025agentsense, yan2025beyond, zhang2024exploring} and multi-robot applications \cite{liu2024leveraging, kannan2024smart, liu2025coherent,  yang2025embodiedbench}. Previous work also exhibited the necessity of proactive communication in cooperation \cite{zhang2024proagent, sun2025assistantx} and studied the feasibility of building LLM agents for collaboration under partial observation as separate agents \cite{li2023theory}. In contrast, our work further studies communication under partial and asymmetric observations for deliberative collaboration and joint decision making. 

\subsection{Goal-Oriented Dialogue}
Our work generally falls in the goal-oriented dialogue system category, where agents converse and act for specific goals, and it has broad application scenarios such as healthcare \cite{valizadeh2022ai} and robotics applications \cite{mandi2024roco, WangRSS24,li2025hmcf, zhang2025towards}.
Compared to classical task-oriented dialogues where agents are designed to serve specific purposes \cite{ xu2024rethinking, ulmer2024bootstrapping,rim2025chat, du2025multi} or that employ a multi-agent system for a specific application \cite{han2024ibsen}, we focus more on the application scenarios that necessitate deliberation between agents to align information and reach an agreement on the final decision. 
Compared to previous works that study negotiation or deliberation agents, where agents need to reach an agreement on an issue, and each agent may have its own goal \cite{xia2024measuring, abdelnabi2024cooperation, karanam2024towards, li2025advancing, hu2025debate}, our work focuses on a pure cooperative scenario where all the agents share the same reward and work towards the same goal.
We regard \cite{lin2024decision} as the closest related work, which also studies collaborative decision-making scenarios. 
While they focus on human-AI collaboration with agents as assistants, our work studies autonomous agent teams with equal status cooperating on human-assigned tasks.

\section{Conclusion}
\vspace{-0.1cm}

In this paper, we formalize deliberative collaboration as a partially observable joint decision-making problem and build a benchmark with a set of tightly related task instantiations with different observation structures and coordination requirements to study how shared decision mechanisms behave under varying assumptions about information availability, tool access, and agent roles.

Through a comprehensive evaluation, we show that current LLM agents remain insufficient for reliable deliberative collaboration. LLM agents may suffer from insufficient information exchange and aggregation, inadequate reasoning capabilities, and limited mathematical capabilities. 
Yet deliberation can also provide opportunities for reflection and error correction.

Overall, our benchmark serves not only as an evaluation framework but also as a diagnostic tool for analyzing where and why deliberative collaboration fails. We hope this work encourages future research on more robust coordination strategies, calibrated tool use, and principled information aggregation for multi-agent LLM systems.

\newpage

\section*{Limitations}
Our benchmark evaluates LLM-agent systems under a fixed reference scaffold. Since the tasks are complex and require multiple capabilities, performance may be sensitive to agent architecture, prompt design, and tool interfaces. Exhaustively evaluating alternative scaffolds is beyond the scope of this work; therefore, our results should be interpreted as representative performance under the evaluated scaffold, rather than a complete characterization of each LLM.

Another limitation is the scale of the evaluation. Due to computational and financial constraints, we evaluate a representative set of models rather than all accessible LLMs, and we do not include extensive sensitivity analyses over partner composition, decoding settings, deliberation budgets, or human baselines. These omissions limit the scope of the empirical conclusions. Future work can instantiate alternative scaffolds, cross-play protocols, communication constraints, or human calibration using the benchmark generation and evaluation code.

\bibliography{custom}

\appendix

\section{Benchmark Details}
\label{sec: benchmark_appendix}

\subsection{Menu Design}
\label{sec: menu_design_appendix}

\subsubsection{Game Formulation}

We implement the menu design game in a basic problem setting: the ingredient state $S^I$ is a vector where $S^I_{i}$ is the amount of the $i$-th ingredient; the cookbook offers a recipe matrix $A$, where $A_{ij}$ denotes the amount of the $i$-th ingredient needed for the $j$-th dish; the user preferences $S^U$ are simplified to a scoring matrix, where $S^U_{kj}$ denotes the score that the $k$-th guest gives to the $j$-th dish; and the agents directly access a part of $S^I$ and $S^U$ as their observation, with the guarantee that $O_{1} \cup O_{2} = S$ and $O_{1} \cap O_{2} = \emptyset$, denoted as \textit{partitioned information}.

By defining a scoring vector $c$ as $c_j = \sum_{k}^{n_{\text{guest}}}S^U_{kj}$ (where $n_{\text{guest}}$ is the number of guests), the final reward is calculated with:
$$ R(d) = \text{IsValid}(d, S^I) \cdot \sum_{j \in d} c_j, $$
where $\text{IsValid}$ is a function indicating whether the menu is valid, specifically, the consumed ingredients should be less than or equal to the available ingredients. The reward function also implies that a dish can only appear once in the final menu.

Additionally, if the agents have access to the full state, the task can be formulated as an integer programming problem:
\begin{align}
    \max_{x \in \{0, 1\}^{n_{\text{dish}}}} c^Tx \notag \\
    s.t. Ax \leq S^I \notag 
\end{align}
Here $n_{\text{dish}}$ is the number of dishes. This implies that the problem has a straightforward solution in which the two agents exchange their observations to get enough information for planning and then solve the integer programming problem. Therefore, the challenge principally lies in choosing the most important information to exchange, expressing and understanding essential information in natural language dialog, and solving the programming problem with intrinsic planning capabilities or utilizing external tools. 
In this work, we use the \textbf{pulp} package as the external solver for the integer program.

\subsubsection{Characters and Preferences}
We define characters as described in the paper.
We present an example below: 
\begin{lstlisting}[style=jsonstyle]
"Name": "Alice",
"Gender": "Female",
"Nationality": "American",
"Age": 30,
"Occupation": {
    "Job Title": "Software Engineer",
    "Industry": "Technology",
    "Work Schedule": "9am - 5pm"
},
"Recent Status": {
    "Current Focus": "Project deadline approaching",
    "Recent Dietary Focus": "Looking for healthy, comforting meals to recharge",
    "Stress Level": "High"
},
"Dietary Preferences": {
    "Vegetarian": false,
    "Vegan": false,
    "Gluten-free": true,
    "Dairy-free": false,
    "Low-carb": false
},
"Food Allergies": ["Nuts", "Dairy"],
"Cultural Preferences": {
    "Cuisine Type": "Italian",
    "Spice Level Preference": "Mild",
    "Traditional Dishes": true
},
"Health and Fitness Goals": {
    "Calorie-conscious": true,
    "Protein-rich": false,
    "Low-sodium": false,
    "Low-sugar": false,
    "Weight-loss focus": false,
    "Muscle-building focus": false
},
"Personality and Dining Style": {
    "Adventurous Eater": false,
    "Comfort Food Lover": true,
    "Indulgent Food Preferences": true,
    "Minimalist Dining": false,
    "Social Eater": true,
    "Snack-friendly": true
},
"Ethical and Religious Considerations": {
    "Halal": false,
    "Kosher": false,
    "Buddhist": false,
    "Ethical Eating Preferences": "Organic"
}
\end{lstlisting}
Based on the persona information, each character is associated with a set of preference values, which are their ratings for each dish.
The value of a dish to the character is defined as $r_{d} (n_{i} + 0.2*n_{i}^{1.5})$, where $r_{d}$ is the rating for that dish and $n_{i}$ is the total number of ingredients required for that dish. We use this formula to encourage the agents to cooperate and choose more complex dishes.

\subsubsection{Agent Observation}
Across menu-design settings, we provide observations on the ingredient inventory in JSON dictionary format. To maintain a balance between agents, each agent will observe at least 40\% and at most 60\% of the ingredients. 

In the \textit{menu-numeric} domain, each agent observes the value of dishes to one guest. In the \textit{menu-semantic} domain, each agent observes partial persona information of both guests, according to the \textit{observed probability} of each field in the person information. The \textit{observed probabilities} are as follows:
\begin{lstlisting}[style=jsonstyle]
"Agent 1": {
    "Name": 1.0,
    "Gender": 1.0,
    "Nationality": 1.0,
    "Age": 1.0,
    "Occupation": 0.7,
    "Recent Status": 0.6,
    "Dietary Preferences": 0.6,
    "Food Allergies": 0.7,
    "Cultural Preferences": 0.6,
    "Health and Fitness Goals": 0.4,
    "Personality and Dining Style": 0.5,
    "Ethical and Religious Considerations": 0.5
},
"Agent 2": {
    "Name": 1.0,
    "Gender": 1.0,
    "Nationality": 1.0,
    "Age": 1.0,
    "Occupation": 0.6,
    "Recent Status": 0.6,
    "Dietary Preferences": 0.7,
    "Food Allergies": 0.8,
    "Cultural Preferences": 0.6,
    "Health and Fitness Goals": 0.5,
    "Personality and Dining Style": 0.7,
    "Ethical and Religious Considerations": 0.6
}
\end{lstlisting}

\subsubsection{Task Selection}
To avoid impractical tasks and ensure the tasks are appropriate to serve as benchmarks for LLM agents, we first randomly generate 10,000 tasks for both \textit{menu-numeric} and \textit{menu-semantic}, and then select the appropriate ones according to the following principles:
\begin{itemize}
    \item There must be at least $3$ dishes in the solved best menu.
    \item There exist 8 kinds of ingredients with non-zero amounts.
    \item Then we select the ones that utilize more ingredients, with a smaller ingredient list, and have more possible dishes and fewer dishes in the best menu. We expect the selected tasks to be feasible yet moderately challenging to find the best menu.

\end{itemize}
For all tasks, there must exist at least 5 possible dishes, the number of possible dishes must account for at least $40\%$ of the total dishes (we sample a subset of dishes for each task).

\subsection{Task Allocation with Asymmetric Roles and Resources} 
\label{sec: task_allocation_appendix}

\subsubsection{Roles and Preferences} 
Distinct from the symmetric peer setting in menu design, we define asymmetric roles for agents in this task. The team consists of three agents: two Workers (Agent 0 and Agent 1) who contribute information and proposals, and one Leader (Agent 2) who holds the exclusive authority to make the final decision to terminate the negotiation.

Each agent is associated with a specific efficiency profile, representing their utility for completing specific tasks. The preference structure is defined as follows: 
\begin{lstlisting}[style=jsonstyle]
{
    "Agent Roles": {
        "agent_0": "Worker",
        "agent_1": "Worker",
        "agent_2": "Leader"
    },
    "Efficiency Profile (overall_preferences)": {
        "agent_0": {
            "task_1": 0.8,
            "task_2": 0.6,
            "task_3": 0.9
        },
        "agent_1": {
            "task_1": 0.7,
            "task_2": 0.8,
            "task_3": 0.5
        },
        "agent_2": {
            "task_1": 0.9,
            "task_2": 0.5,
            "task_3": 0.7
        }
    }
}
\end{lstlisting}

The team utility is calculated as the sum of efficiencies for the assigned tasks. Formally, the reward function is defined as $R(x) = \sum_{i=0}^{N-1} \sum_{j=0}^{M-1} v_{i,j} \cdot x_{i,j}$, where $v_{i,j}$ is the efficiency of agent $i$ for task $j$, and $x_{i,j} \in \{0,1\}$ indicates whether task $j$ is assigned to agent $i$.

\subsubsection{Agent Observation}
In this setting, observations are structured around resource availability. The ground truth state $S$ comprises private resources $S^{priv}$ (exclusive to specific agents, e.g., Time, GPU) and public resources $S^{pub}$ (shared across the team, e.g., Budget).

The observation space is partially partitioned and additive. Specifically:
\begin{enumerate}
    \item Private Resources: Each agent fully observes their own private resources, which are hidden from others.
    \item Public Resources: Each agent observes only a fragment of the shared public resources. The system guarantees that the sum of all agents' observed public resources equals the ground truth total.
\end{enumerate}

An example of a single agent's observation (e.g., Agent~0) is shown below:
\begin{lstlisting}[style=jsonstyle]
{
    "current_private_resources": {
        "Time": 10,  // Fully observed, exclusive to agent_0
        "GPU": 2
    },
    "current_public_resources": {
        "Budget": 80 // Partial fragment; total might be 200
    },
    "efficiency_dict": {
        "task_1": 0.8, // Self-efficiency is known
        "task_2": 0.6,
        "task_3": 0.9
    },
    "task_requirements": {
        "task_1": {
            "Time": 3, 
            "GPU": 1, 
            "Budget": 100
        },
        "task_2": {
            "Time": 4, 
            "GPU": 2, 
            "Budget": 50
        }
    }
}
\end{lstlisting}

Agents must communicate to aggregate the partial public observations into a global view to verify if a proposed allocation satisfies the global public resource constraints.

\subsubsection{Task Selection}
To avoid impractical tasks and ensure the scenarios provide appropriate challenges for hierarchical coordination, we first randomly generate 1,000 candidate instances for each domain by sampling tasks, agent personas, and resource capacities, and then select the appropriate ones according to the following principles:

\begin{itemize}
    \item \textbf{Feasibility Constraint}: There must be at least $3$ tasks successfully assigned in the optimal solution ($n_{done} \geq 3$). This filters out instances that are unsolvable or too trivial due to extreme resource shortages.
    \item \textbf{Quality Prioritization}: We score candidates using a priority function defined as $P = w_{p} \cdot p_{done} + w_{u} \cdot U_{res} + w_{r} \cdot R_{max}$, where $p_{done}$ is the task completion ratio and $U_{res}$ is the average resource utilization. We select instances that maximize these metrics to ensure the scenarios allow for high task throughput while maintaining tight resource constraints, requiring efficient planning rather than having overabundant resources.
\end{itemize}

\section{Experiments}
\label{sec: appendix_exp}

This section presents additional experimental results and metric details.

\begin{table*}[t]
\centering
\setlength{\tabcolsep}{1.75pt}
\begin{tabular}{l c cc cc cc}
\hline
\multirow{2}{*}{Model} &
\multirow{2}{*}{Tools} &
\multicolumn{2}{c}{Menu-Numeric} &
\multicolumn{2}{c}{Menu-Semantic} &
\multicolumn{2}{c}{Task Allocation} \\
\cline{3-8}
& & NAR (mean$\pm$se) & VR(\%) & NAR (mean$\pm$se) & VR(\%) & NAR (mean$\pm$se) & VR(\%) \\
\hline
\multicolumn{8}{c}{\textit{Deliberative Collaboration}} \\
\hline

GPT-5.1 & \ding{55} & 95.60$\pm$1.51 & 100.00 & 79.91$\pm$3.74 & 60.00 & 80.77$\pm$2.55 & 98.33 \\
        & \ding{51} & 90.72$\pm$2.83 & 98.33  & 72.48$\pm$3.97 & 66.67 & 76.98$\pm$2.79 & 96.67 \\

DeepSeek-V3.2 & \ding{55} & 91.51$\pm$2.51 & 91.67 & 81.13$\pm$3.37 & 93.33 & 70.62$\pm$3.02 & 61.67 \\
              & \ding{51} & 96.65$\pm$1.22 & 98.33 & 83.90$\pm$2.34 & 100.00 & 85.51$\pm$2.52 & 90.00 \\

GLM-4.7 & \ding{55} & 92.72$\pm$1.81 & 98.33 & 66.99$\pm$3.31 & 95.00 & 72.21$\pm$2.81 & 73.33 \\
        & \ding{51} & 91.18$\pm$2.26 & 93.33 & 67.94$\pm$3.06 & 93.33 & 73.51$\pm$2.26 & 83.33 \\

GPT-4.1-mini & \ding{55} & 41.35$\pm$3.80 & 13.33 & 54.70$\pm$4.19 & 35.00 & 64.44$\pm$3.01 & 65.00 \\
             & \ding{51} & 65.21$\pm$4.10 & 38.33 & 59.84$\pm$3.99 & 30.00 & 80.29$\pm$2.13 & 80.00 \\

Qwen3-Next-80B & \ding{55} & 60.41$\pm$4.24 & 41.67 & 48.60$\pm$3.87 & 75.00 & 62.07$\pm$2.22 & 83.33 \\
               & \ding{51} & 57.96$\pm$4.21 & 28.33 & 58.59$\pm$3.87 & 76.67 & 60.16$\pm$2.33 & 83.33 \\

Qwen3-32B & \ding{55} & 58.73$\pm$3.58 & 6.67  & 53.55$\pm$3.07 & 18.33 & 40.44$\pm$1.84 & 11.67 \\
          & \ding{51} & 56.90$\pm$4.29 & 25.00 & 48.92$\pm$3.44 & 28.33 & 74.35$\pm$2.96 & 55.00 \\

Qwen3-30B & \ding{55} & 51.80$\pm$4.39 & 36.67 & 51.08$\pm$3.30 & 50.00 & 60.04$\pm$1.91 & 90.00 \\
          & \ding{51} & 51.65$\pm$4.24 & 36.67 & 52.08$\pm$3.51 & 53.33 & 64.50$\pm$2.33 & 91.67 \\

\hline
\multicolumn{8}{c}{\textit{Centralized Baseline}} \\
\hline

GPT-5.1 & \ding{55} & 96.71$\pm$1.22 & 100.00 & 90.67$\pm$1.37 & 100.00 & 91.67$\pm$0.95 & 100.00 \\
        & \ding{51} & 98.42$\pm$0.80 & 100.00 & 90.84$\pm$1.46 & 100.00 & 93.96$\pm$1.44 & 98.33 \\

DeepSeek-V3.2 & \ding{55} & 97.36$\pm$1.00 & 100.00 & 92.46$\pm$1.35 & 100.00 & 85.01$\pm$2.24 & 88.33 \\
              & \ding{51} & 98.14$\pm$0.84 & 100.00 & 92.28$\pm$1.21 & 100.00 & 94.59$\pm$1.90 & 90.00 \\

GLM-4.7 & \ding{55} & 97.19$\pm$0.96 & 98.33 & 81.41$\pm$2.20 & 93.33 & 72.73$\pm$3.08 & 66.67 \\
        & \ding{51} & 98.81$\pm$0.67 & 98.33 & 87.70$\pm$1.94 & 98.33 & 79.02$\pm$3.00 & 70.00 \\

GPT-4.1-mini & \ding{55} & 94.64$\pm$1.25 & 100.00 & 85.55$\pm$2.10 & 98.33 & 57.26$\pm$2.80 & 56.67 \\
             & \ding{51} & 99.26$\pm$0.43 & 100.00 & 87.88$\pm$1.72 & 100.00 & 70.07$\pm$3.27 & 48.33 \\

Qwen3-Next-80B & \ding{55} & 93.81$\pm$1.62 & 95.00 & 80.87$\pm$2.16 & 96.67 & 83.44$\pm$2.27 & 86.67 \\
               & \ding{51} & 98.58$\pm$0.73 & 96.67 & 81.96$\pm$2.15 & 98.33 & 83.42$\pm$2.74 & 73.33 \\

Qwen3-32B & \ding{55} & 83.72$\pm$2.46 & 13.33 & 72.69$\pm$2.55 & 21.67 & 37.27$\pm$1.68 & 1.67 \\
          & \ding{51} & 98.06$\pm$0.71 & 91.67 & 86.99$\pm$2.00 & 76.67 & 69.68$\pm$3.04 & 28.33 \\

Qwen3-30B & \ding{55} & 97.46$\pm$1.20 & 93.33 & 82.94$\pm$2.10 & 98.33 & 73.70$\pm$2.69 & 66.67 \\
          & \ding{51} & 98.45$\pm$0.71 & 100.00 & 78.82$\pm$2.30 & 98.33 & 70.54$\pm$2.86 & 40.00 \\

\hline
\end{tabular}

\caption{Overall results across domains using normalized adjusted reward (NAR, mean$\pm$standard error) and valid ratio (\%).}
\label{tab: append_results_nar}
\end{table*}

\subsection{Metric Detail}
\label{sec: appendix_Metric_Details}

In this section, we detail the evaluation metrics used to assess the performance of the agents. These metrics evaluate the validity, optimality, and robustness of the joint decisions, as well as the accuracy of the agents' internal state estimation.

\subsubsection{Validity of Decision}
A joint decision is considered valid if and only if it satisfies all domain-specific feasibility constraints defined in the environment. These typically include \textit{capacity constraints} (e.g., ensuring resource consumption does not exceed availability) and \textit{consistency constraints} (e.g., ensuring logical assignment rules are respected). Only decisions that fully adhere to these hard constraints are eligible for positive rewards.

\subsubsection{Normalized Reward (NR)}
Normalized Reward (NR) measures the optimality of the final decision relative to the theoretical upper bound. It is defined as:
\begin{equation}
    \text{NR} = \frac{R_{final}}{R_{max}} \times 100
\end{equation}
where $R_{final}$ is the reward achieved by the agents' final proposal, and $R_{max}$ is the maximum possible reward computed by the integer programming solver under the ground truth state. If the decision violates any feasibility constraint, $R_{final}$ is set to 0.

\subsubsection{Normalized Adjusted Reward (NAR)}
To provide a more fine-grained assessment for invalid outcomes, we introduce Normalized Adjusted Reward (NAR), which credits partial success while penalizing infeasibility. It is calculated as:
\begin{equation}
    \text{NAR} = 
    \begin{cases} 
    \text{NR} & \text{valid} \\
    \frac{R_{subset}}{R_{max}} \times \frac{|D_{subset}|}{|D_{original}|} \times 100 & \text{invalid}
    \end{cases}
\end{equation}
For invalid proposals, we employ a greedy algorithm to extract a \textbf{maximal feasible subset} of the decision components (e.g., a subset of valid assignments or items) from the original proposal. Here, $|D_{subset}|$ and $|D_{original}|$ represent the size of the valid subset and the original proposal, respectively. This metric evaluates the agents' ability to identify valid sub-solutions even when the global plan is flawed.

\subsubsection{Valid Ratio (VR)}
Valid Ratio (VR) measures the system's reliability in adhering to constraints. It is defined as the percentage of evaluation instances where the agents successfully reach a fully valid final decision:
\begin{equation}
    \text{VR} = \frac{N_{valid}}{N_{total}} \times 100
\end{equation}
where $N_{valid}$ is the number of sessions concluding with a valid outcome ($R_{final} > 0$).

\subsubsection{Hallucination Rate (HR)}
Hallucination Rate (HR) measures how often agents overestimate the availability of resources or ingredients. A state component $i$ is considered "hallucinated" if the agent's estimated quantity $S^{Est}_i$ strictly exceeds the ground truth $S^{GT}_i$. The metric is defined as the percentage of such overestimated components across the entire state vector:
\begin{equation}
    \text{HR} = \frac{\sum_{i=1}^{D} \mathbb{I}(S^{Est}_i > S^{GT}_i)}{D}
\end{equation}
where $D$ represents the total dimension of the state space, and $\mathbb{I}(\cdot)$ is the indicator function. By tracking these "false positives" in resource availability, HR assesses the risk of agents proposing infeasible plans based on non-existent resources.

\subsubsection{Normalized Mean Absolute Error (NMAE)}
For task allocation, hallucination rate is often close to zero and therefore does not capture errors in aggregating pooled public resources. We additionally measure normalized mean absolute error (NMAE) over public resources as a state-estimation error metric. Let $K_{pub}$ be the union of public-resource keys in the estimated and ground-truth dictionaries:
\begin{equation}
    \text{NMAE}_{pub}
    = \frac{1}{|K_{pub}|}\sum_{k\in K_{pub}} e_k,\quad
    e_k=\frac{|\hat{v}_k-v_k|}{|v_k|}
\end{equation}
for nonzero $v_k$, with zero ground-truth dimensions assigned error $1.0$ if $\hat{v}_k\neq 0$ and $0.0$ otherwise.

\begin{table}[t]
\centering
\setlength{\tabcolsep}{2pt}
\resizebox{\linewidth}{!}{
\begin{tabular}{lccc}
\hline
Model & Menu-Numeric & Menu-Semantic & Task Allocation \\
\hline
GPT-5.1 & $0.36\pm0.21$ & $0.47\pm0.26$ & $0.00\pm0.01$ \\
DeepSeek-V3.2 & $0.38\pm0.18$ & $0.38\pm0.20$ & $0.00\pm0.01$ \\
GLM-4.7 & $0.37\pm0.19$ & $0.31\pm0.18$ & $0.00\pm0.01$ \\
GPT-4.1-mini & $0.56\pm0.24$ & $0.51\pm0.22$ & $0.00\pm0.01$ \\
Qwen3-Next-80B & $0.49\pm0.21$ & $0.28\pm0.19$ & $0.00\pm0.00$ \\
Qwen3-32B & $0.43\pm0.20$ & $0.33\pm0.18$ & $0.00\pm0.01$ \\
Qwen3-30B & $0.42\pm0.20$ & $0.39\pm0.20$ & $0.00\pm0.00$ \\
\hline
\end{tabular}
}
\caption{Hallucination rate (HR, mean $\pm$ standard deviation) aggregated over non-oracle settings.}
\label{tab:hr}
\end{table}

\begin{table}[t]
\centering
\setlength{\tabcolsep}{4pt}
\resizebox{\linewidth}{!}{
\begin{tabular}{lccc}
\toprule
Domain & $\beta$ (HR $\rightarrow \Delta_{\text{cent}}$) & $R^2$ & $p$ \\
\midrule
Menu-Numeric     & $61.35$  & 0.0775 & $<.001$ \\
Menu-Semantic    & $37.73$  & 0.0373 & $<.001$ \\
Task Allocation  & $277.43$ & 0.0014 & $0.272$ \\
Overall          & $67.56$  & 0.1300 & $<.001$ \\
\bottomrule
\end{tabular}
}
\caption{
Linear regression results relating hallucination rate (HR) to the centralized performance gap.
$\beta$ denotes the regression coefficient of HR, $R^2$ denotes the proportion of variance in the gap explained by HR, and $p$ denotes the p-value for the correlation test.
}
\label{tab: hr_centralized_gap}
\end{table}

\subsection{Supplementary Results}
\label{sec: appendix_Supplementary_results}

\subsubsection{Model Performance Measured with NAR}
To measure the subtle differences between model preferences and provide diagnostic indicators to study model preferences and characteristics, we report results using NAR in \Cref{tab: append_results_nar}.
In contrast to NR, which strictly punishes invalid plans, NAR is a much more tolerant indicator, where small models with low VR can still obtain a modest reward; for example, Qwen3-32B achieves an average NAR of 58.73 in the \textit{menu-numeric} settings, but only an average NR of 4.78 as reported in \Cref{tab: main_results}.

\subsubsection{State Estimation Error Diagnostics}
\label{sec: appendix_hr_results}

To inspect information exchange and aggregation during deliberation, we use HR and $\text{NMAE}_{pub}$ as complementary diagnostics of state-estimation error. 
As shown in \Cref{tab:hr}, agents sometimes overestimate unavailable resources in the menu-design domains, whereas HR is near zero in task allocation. This suggests that HR captures menu-design hallucination but is less informative for task allocation, where errors mainly involve aggregating pooled public resources.

We then examine whether these state-estimation errors are associated with performance gaps. For menu design, linear regression in \Cref{tab: hr_centralized_gap} shows a significantly positive correlation between HR and $\Delta_{\text{cent}}$, suggesting that hallucinated resources account for part of the decentralized performance gap. This is consistent with the view that centralized access reduces errors from imperfect information aggregation.
In task allocation, because HR is near zero, we instead analyze public-resource aggregation error using $\text{NMAE}_{pub}$. In the tool-augmented subset, higher $\text{NMAE}_{pub}$ is significantly associated with lower NR (Pearson $r=-0.2513$, $p<.0001$; Spearman $\rho=-0.2886$, $p<.0001$), while the association is not statistically significant in the no-tools condition.

\section{Case Studies}
\label{sec:appendix_case_studies}

In the main text, we report aggregate performance trends. This section presents some case studies that illustrate the underlying mechanisms behind these phenomena.

\subsection{A Failed Case of the Centralized Baseline}
\label{sec:appendix_oracle_case_study}

\begin{table}[t] 
\centering
\hrule height 0.8pt \vspace{3pt}
\textbf{Prompt: Failed Centralized Baseline Example} \vspace{3pt}
\hrule height 0.4pt 

\begin{lstlisting}[style=smallpromptstyle]
== BASELINE EXPERIMENT ==
You have access to ALL information from all agents and must make the final
allocation directly (no negotiation).

== SCENARIO ==
Allocate 10 tasks among 3 agents to maximize team value under resource limits.
Use the given task names as JSON keys.

== COMPLETE INFORMATION ==
<agent_0_efficiency_values>
{full efficiency dictionary for agent_0}
</agent_0_efficiency_values>
<agent_1_efficiency_values>
{full efficiency dictionary for agent_1}
</agent_1_efficiency_values>

== TOTAL RESOURCES ==
CRITICAL:
- Private resources are per-agent (do NOT add across agents).
- Public resources MUST be pooled by summation across all observations.

<total_private_resources>
Agent 0: {"Time": 6.0, "Mental": 4.6, "Physical": 3.0, "Multitasking": 1.8}
</total_private_resources>

<task_requirements>
{full resource requirements for 10 tasks; omitted}
</task_requirements>

== YOUR TASK ==
Produce a valid allocation JSON and explicitly verify:
(1) each agent's private constraints;
(2) pooled public constraints.
Output format: 
<reasoning_process> ... </reasoning_process>
<decision>{... "final_decision": {...}}</decision>
\end{lstlisting}
\caption{The prompt for the centralized baseline in a failed example.}
\label{tab:appendix-failed-cent}
\end{table}

We present a case study contrasting the centralized baseline (full information, no deliberation) with the multi-agent deliberative system on the same task-allocation instance. The prompt used in this failed example is shown in \Cref{tab:appendix-failed-cent}. 

Although the centralized baseline receives complete information, it fails to produce a valid final allocation. It correctly identifies that assigning \texttt{Prep vegetables batch} to Agent~0 violates the time limit, but still assigns that task to Agent~0 in the \texttt{final\_decision}, contradicting its own earlier validation.

\begin{lstlisting}[style=textblockstyle]
[Earlier validation]
Agent 0: Store ingredients + Receive delivery + Prep vegetables
Time = 2 + 2 + 3 = 7 > 6  (NOT OK)

[Later final JSON]
"Store ingredients with FIFO labeling": "agent_0",
"Receive delivery and verify items": "agent_0",
"Prep vegetables batch (wash/chop/portion)": "agent_0"   <-- invalid
\end{lstlisting}

By contrast, the multi-agent system can perform a distributed verification, whereby each agent examines whether the proposal violates the constraints visible from its local perspective, thereby preventing invalid plans from being finalized.

\begin{lstlisting}[style=textblockstyle]
Round 2 (agent_0): I cannot accept agent_2's proposal because it exceeds
their private resource limits significantly...

Round 2 (agent_2): I cannot accept the current partner proposal because it
overloads agent_1 in time/mental/physical resources and exceeds serving and
kitchen limits...
\end{lstlisting}

\subsection{An Example of the ``Skepticism Trap'' in Tool-Augmented Reasoning}
\label{sec:appendix_skepticism_trap_case_study}

\begin{table}[t] 
\centering
\hrule height 0.8pt \vspace{3pt}
\textbf{Prompt: ``Skepticism Trap'' Input Example} \vspace{3pt}
\hrule height 0.4pt 

\begin{lstlisting}[style=smallpromptstyle]
You are {agent_name} in a 3-agent team (2 workers + 1 leader).
Your goal is to maximize total team value under private + public constraints.

IMPORTANT environment rule:
- Private resources are agent-specific.
- Public resources are pooled across agents by SUMMATION after information sharing.

Your local observation:
<current_private_resources>
{...}
</current_private_resources>
<current_public_resources>
{partial fragment only}
</current_public_resources>
<efficiency_dict>
{...}
</efficiency_dict>
<task_requirements>
{...}
</task_requirements>

(If tools are enabled)
<tool_output>
best_allocation = solver result computed from your CURRENT estimated totals
</tool_output>

Deliberation protocol:
- Share observations and challenge infeasible proposals.
- The Leader (agent_2) makes the final decision.
- Output proposal/decision in the required XML+JSON format.
\end{lstlisting}
\caption{The prompt for the decision module with external tools in an example of the ``skepticism trap''.}
\label{tab:appendix-failed-tool}
\end{table}

We present a case study that illustrates a counterintuitive result from the task-allocation domain: the tool-augmented setting fails to produce a valid allocation, whereas the no-tool setting converges to a valid conservative plan. The prompt structure for this scenario is shown in \Cref{tab:appendix-failed-tool}.

The key issue in this failed example is a misinterpretation of tool context, which triggers a cascade of over-correction. The agents observe different fragments of public resources; globally, the public constraint is the sum across agents. However, some agents misinterpret their local fragment as the team-level hard cap and consequently judge the tool-generated solution as infeasible, since it exceeds the resources visible from their local view alone. In this case, the agents not only need to think about the proposal but also need to verify the suggested decision, adding an extra burden. We show the summarized example output below:

\begin{lstlisting}[style=textblockstyle]
Round 1 (agent_0):
I see Ingredients=63, Serving=2, Kitchen=2, Sanitation=1.
The "best_allocation" uses 200+ ingredients and much more kitchen/sanitation,
so I'm worried it is not feasible.

Round 1 (agent_1):
I also see very small public resources (Ingredients=74, Serving=2, Kitchen=2,
Sanitation=1). I don't trust the earlier tool-based allocation.

Round 1 (agent_2, Leader):
I see a much larger fragment (Ingredients=204, Serving=4, Kitchen=4,
Sanitation=2), but I will align with the team's concern and move to a safer plan.
\end{lstlisting}

The true global public budget is the \emph{sum} of fragments (e.g., Ingredients \(=63+74+204=341\)), so the tool output was feasible under the true pooled budget.

By contrast, in the no-tool condition, agents still do not perfectly reconstruct the pooled public totals, but they are no longer distracted by a proposal that appears incompatible with local views.
The dialogue focuses on progressive synchronization and converges to a valid conservative allocation.

\begin{lstlisting}[style=textblockstyle]
Round 2 (agent_2, Leader):
Let's optimize for worst-case robustness under the tightest plausible public
limits (Ingredients 63, Serving 2, Kitchen 2, Sanitation 1) and finalize a
minimal safe bundle.

[Later rounds]
All agents confirm the same small allocation and keep it consistent through the
final decision output.
\end{lstlisting}

\section{Prompts}
\label{sec:agent_appendix}

Since the original prompts are too long to include in full, we provide the general organizational format of the prompts (omitting some details). Full prompt templates can be found in our code. 
The prompts used in each task setting are generally consistent, with minor specialization on instructions.

In this appendix, we show the prompt templates in the menu design domain as an example, including the observation module in \Cref{tab: obs_prompt}, the planning module in \Cref{tab: planning_prompt}, the decision module in \Cref{tab: dec_prompt}, and the conversation module in \Cref{tab: conv_prompt}.

\begin{table*}[htbp]

\centering
\hrule height 0.8pt \vspace{3pt}
\textbf{Prompt Template: Observation Module} \vspace{3pt}
\hrule height 0.4pt 

\begin{lstlisting}[style=promptstyle]
You are an AI agent, named {agent_name}. The user asked you and another agent to design a menu for treating their guests. ... 
(setting-specific content)
Menu-Numeric: The preferences, at this stage, are numerical scores, ...
Menu-Semantic: ... You need to estimate the preferences for each dish based on the information of both guests. ...

...You may need to communicate for collaboration. You can communicate for {max_round} rounds. In each round, you can say up to {max_character} characters, messages that exceed this limit will be truncated. ...

Here you will do the first step, you need to estimate the "ingredients" and the "preference". ...

Your initial observation: {current_obs} {score_dict}
The recipes are: {recipes}
The dialog is: {chat_history} 
...
Format:

<thinking_process_1>
thinking_process...
</thinking_process_1>

<partner_ingredients>
{{ "ingredient_1":1,...}}
</partner_ingredients>

<partner_preference>
{{"dish_1":1.5,...}}
</partner_preference>

<thinking_process_2>
thinking_process...
</thinking_process_2>

<total_ingredients>
{{"ingredient_1":1,...}}
</total_ingredients>

<overall_preference>
{{ "dish_1":1.5,...}}
</overall_preference>
\end{lstlisting}

\caption{Simplified prompt template for the observation module.}
\label{tab: obs_prompt}
\end{table*}

\begin{table*}[htbp]
\centering
\hrule height 0.8pt \vspace{3pt}
\textbf{Prompt Template: Planning Module} \vspace{3pt}
\hrule height 0.4pt 

\begin{lstlisting}[style=promptstyle]
You are an AI agent, named {agent_name}. The user asked you and another agent to design a menu for treating their guests. ... 

(setting-specific content)
Menu-Numeric: The preferences, at this stage, are numerical scores, ...
Menu-Semantic: ... You need to estimate the preferences for each dish based on the information of both guests. ...

...You may need to communicate for collaboration. You can communicate for {max_round} rounds. In each round, you can say up to {max_character} characters, ...

Here you will do the second step, you will need to design the best menu, ...
Now is round {round_count} of the total {max_round} rounds.

Here is your current estimation: {total_ingredients}, {overall_preference}
Here is the recipes that indicate the ingredients need by each dish: {recipes}

(if use tools )
Here is a list of dishes which already have sufficient ingredients: {available_dishes}
Here is a list of missing ingredients for currently unavailable dishes: {unavailable_dishes_info}
Here is the optimal dish list based on total ingredients and overall preference: {best_menu}

...

Format:
<reasoning_process>
XX
</reasoning_process>

<proposal>
{{"menu_proposal": ["dish_1","dish_2","dish_3","dish_4", ...],
  "explanation": "XX"}}
</proposal>
\end{lstlisting}
\caption{Simplified prompt template for the planning module.}
\label{tab: planning_prompt}
\end{table*}

\begin{table*}[htbp]
\centering
\hrule height 0.8pt \vspace{3pt}
\textbf{Prompt Template: Decision Module} \vspace{3pt}
\hrule height 0.4pt 

\begin{lstlisting}[style=promptstyle]
You are an AI agent, named {agent_name}. The user asked you and another agent to design a menu for treating their guests. ... 

(setting-specific content)
Menu-Numeric: The preferences, at this stage, are numerical scores, ...
Menu-Semantic: ... You need to estimate the preferences for each dish based on the information of both guests. ...

...You may need to communicate for collaboration. You can communicate for {max_round} rounds. In each round, you can say up to {max_character} characters, ...

Here you will do the third step, you need to decide accept your partner's proposal or not....
You have the following information:

Now is round {round_count} of the total {max_round} rounds.
Here is the recipes that indicate the ingredients need by each dish: {recipes}
Total ingredients and the overall preference: {total_ingredients},{overall_preference}

(if use tools )
Based on your estimation:
Here is a list of dishes which already have sufficient ingredients: {available_dishes}
Here is a list of missing ingredients for currently unavailable dishes: {unavailable_dishes_info}
Here is the optimal dish list based on total ingredients and overall preference: {best_menu}

The dialog between you and your partner so far is: {chat_history}
Your latest proposal is: {agent_proposal}
Your explanation of your proposal is:{agent_proposal_explanation}
The latest proposal from your partner is:{partner_proposal},{partner_explanation}

Now you need to decide whether to accept your partner's proposal or not.
...
Format:
<reasoning_process>
XX
</reasoning_process>

<decision>
{{ "accept_decision": true/false, "explanation": "XX"}}
</decision>
\end{lstlisting}
\caption{Simplified prompt template for the decision module.}
\label{tab: dec_prompt}
\end{table*}

\begin{table*}[htbp]
\centering
\hrule height 0.8pt \vspace{3pt}
\textbf{Prompt Template: Conversation Module} \vspace{3pt}
\hrule height 0.4pt 

\begin{lstlisting}[style=promptstyle]
You are an AI agent, named {agent_name}. The user asked you and another agent to design a menu for treating their guests. ... 

(setting-specific content)
Menu-Numeric: The preferences, at this stage, are numerical scores, ...
Menu-Semantic: ... You need to estimate the preferences for each dish based on the information of both guests. ...

...You may need to communicate for collaboration. You can communicate for {max_round} rounds. In each round, you can say up to {max_character} characters, ...

Here you will do the fourth step, also the final step in this round....

Here is all the information so far:

Now is round {round_count} of the total {max_round} rounds.
Here is the recipes that indicate the ingredients need by each dish:{recipes}
Your initial observation:{current_obs},{person_1_info},{person_2_info}
The dialog is:{chat_history}
Your estimation of your partner's observation and the total ingredients and the overall preference: {partner_ingredients}, {partner_preference}, {total_ingredients}, {overall_preference}

(if use tools )
Based on your estimation:
Here is a list of dishes which already have sufficient ingredients: {available_dishes}
Here is a list of missing ingredients for currently unavailable dishes: {unavailable_dishes_info}
Here is the optimal dish list based on total ingredients and overall preference: {best_menu}

Based on these, your latest proposal is: {agent_proposal}
The latest proposal from your partner is:{partner_proposal}
You rejected your partner's proposal, because:{decision_explanation}
...
Now, for the next round, please organize what to talk with your partner.
...
Format:
<reasoning_process>
XX
</reasoning_process>

<message_content>
(The words you want to say to your partner)
</message_content>
\end{lstlisting}
\caption{Simplified prompt template for the conversation module.}
\label{tab: conv_prompt}
\end{table*}

\end{document}